# Human Eye Visual Hyperacuity: A New Paradigm for Sensing?

Adur Lagunas, Oier Domínguez, Susana Martinez-Conde, Stephen L. Macknik, and Carlos del-Río,
*Senior Member, IEEE*

*Abstract*—The human eye appears to be using a low number of sensors for image capturing. Furthermore, regarding the physical dimensions of cones–photoreceptors responsible for the sharp central vision–, we may realize that these sensors are of a relatively small size and area. Nonetheless, the eye is capable to obtain high resolution images due to visual hyperacuity and presents an impressive sensitivity and dynamic range when set against conventional digital cameras of similar characteristics. This article is based on the hypothesis that the human eye may be benefiting from diffraction to improve both image resolution and acquisition process.

The developed method intends to explain and simulate using MATLAB software the visual hyperacuity: the introduction of a controlled diffraction pattern at an initial stage, enables the use of a reduced number of sensors for capturing the image and makes possible a subsequent processing to improve the final image resolution. The results have been compared with the outcome of an equivalent system but in absence of diffraction, achieving promising results. The main conclusion of this work is that diffraction could be helpful for capturing images or signals when a small number of sensors available, which is far from being a resolution-limiting factor.

*Index Terms*—Diffraction, resolution, sensor, visual hyperacuity.

## I. INTRODUCTION

NOWADAYS, embedded digital cameras can be found almost in every smartphone or electronic device, thus making possible to sense or to capture images anywhere and at any time, so it seems not to be a complicated task. Essentially, nearly all conventional cameras work with CCD or CMOS sensors of a certain number of pixels (generally expressed in Megapixels), which define the maximum size and resolution of captured images. Even if the lens system determines to a large extent the final image and camera quality, the feature commonly considered to evaluate digital cameras is the resolution of the sensor: the higher the number of pixels, the higher the performance. Nevertheless, there are many other essential parameters to adequately characterize digital cameras and sensors: sensitivity, dynamic range, the depth of field, sensor size, optics quality, etc.

We may consider the human eye working principle analogous to the optical system of a common digital camera, since the incoming light passes through an aperture (pupil) controlled by a diaphragm (iris), a lens system (crystalline) and finally projecting over the retina where light sensitive vision cells or sensors (cones and rods) are located. These sensors receive information modulated at optical frequencies, which is later introduced into the neural system to be processed by the brain.

It is widely known that the human eye performs impressive tasks related to sensitivity when illumination conditions are not optimal, or the high-resolution details it can obtain, just to mention a few. When analyzing the photoreceptors over the retina, it is easy to realize that sensors used for central sharp vision are quite small in size and relatively few in number. If we consider the human eye operating principle equivalent to that of digital cameras, that would be emulated as a system of low sensitivity and poor resolution. However, it can by far overcome the sensitivity and image resolution defined by the size and number of its sensors thanks to an amazing feature: the visual hyperacuity. Given this fact, it seems logical to think that the human eye goes beyond the working principles of common digital cameras and should be given further consideration.

## II. HUMAN EYE VS. DIGITAL CAMERA SENSORS

The European Machine Vision Association (EMVA), proposed the EMVA Standard 1288 [1] which defines a unified method to measure and compute specification parameters and characterization data for conventional digital cameras or image sensors. The company Point Grey compared multiple color single lens camera models with CCD and CMOS sensors, adopting the EMVA 1288 standard for testing, and published the latest results in the Q2 2016 Color Camera Sensor Review [2]. This document analyzes multiple camera parameters as the quantum efficiency (capacity to convert incoming light into an electrical signal), dynamic range, saturation capacity, temporal dark noise and the absolute sensitivity threshold.

According to the EMVA 1288 Standard, the number of photons received by a pixel or sensor is proportional to its area $A$, the exposure time $t_{exp}$ and the irradiance $E$ on the sensor surface area in W/m$^2$, as follows:

$$\mu_p[\text{photons}] = \frac{AEt_{\exp}}{h\nu} = \frac{AEt_{\exp}}{h\,c/\lambda} \qquad (1)$$

using the quantification of the energy of electromagnetic radiation in units of $h \cdot \nu$, where $h$=6.6260755$\cdot$10$^{-34}$ J$\cdot$s is the Planck constant and $\nu$=$c/\lambda$ the frequency, related to the wavelength $\lambda$, and the speed of light $c$=2.99792458$\cdot$10$^8$ m/s.

Supposing the same irradiance $E$ and exposure time $t_{exp}$ when comparing different sensors, the only factor affecting the number of photons captured by a pixel is the area $A$ of the sensor itself. In the particular case of the human eye, the average size of photoreceptors or cones in the fovea–the area over the retina that deals with sharp central vision–is about 1.5 µm diameter [3], which is quite small in comparison with the pixel sizes of conventional cameras shown in the review (from 2.2 to 9.9 µm). Comparing the area of a cone, assuming circular shape (1.77 µm$^2$), to the area of the smallest pixel size of a conventional camera (4.84 µm$^2$) extracted from [2], the area of a human eye photoreceptor would be 2.74 times smaller, reducing by the same factor the photon flux captured by each sensor.

This could also limit the dynamic range, if we define it as in (2), since saturation capacity is a magnitude related to the pixel size,

$$DR = \frac{u_{p.sat}}{u_{p.\min}} \qquad (2)$$

where $u_{p.sat}$ is the saturation irradiation and $u_{p.\min}$ represents the minimum detectable irradiation or absolute sensitivity threshold. However, the human eye presents an extraordinary dynamic range and detects subtle contrast variations [4], being able to perceive information from very low-light conditions up to the brightness of daylight. Different sources [5], [6], define a detectable luminance range of cone cells for photopic vision (chromatic or color perception) of 10$^8$ or 160 dB when expressed on a logarithmic scale as it is shown in (3).

$$DR_{dB} = 20\log_{10}(DR) \qquad (3)$$

This value largely outperforms the multiple dynamic ranges observed in the camera review, presenting values from 53 to 73 dB approximately, even if the sensors in the human eye are considerably smaller in area. In scientific imaging applications, CMOS sensors with dynamic ranges up to 120 dB can be found, which are achieved by means of different High Dynamic Range (HDR) imaging techniques [7]–[9]. Nonetheless, this values are nowhere near to the ones obtained by the human eye.

Analyzing the maximum image resolution of the human eye in relation to the physical characteristics and distribution of cones over the retina, the fovea centralis is where the maximum density of cones is observed (147.000/mm$^2$ in the central part [10]) being the responsible for color vision and highest visual acuity. The fovea has an approximated diameter of 1.5 mm (an area of 1.767 mm$^2$) [11], what leads us to deduce that the maximum photoreceptors responsible for sharp central vision cannot be more than 260,000 (i.e., 0.26 megapixels).

Even though the human eye has a considerably low number of a reduced size of sensors, the visual system shows an impressive performance with a very robust and simple architecture: a wide dynamic range, high sensitivity to see objects under low-light conditions (controlling the incoming light to prevent saturation), but in this regard, the most relevant feature is the high-resolution images it can obtain. This quality is known as visual hyperacuity, defined as the capability to see beyond the understandable acuity defined by the number and size of photoreceptors. This feature allows the human vision to outperform artificial imaging systems with similar optical and sensor characteristics.

## III. HUMAN EYE VISUAL HYPERACUITY

The retina of the human eye is a light-sensitive layer composed of photoreceptors –rods and cones– and a strongly interconnected biological neural network that communicates with the deeper layers of the brain by means of the optical nerve [12]. Rods are responsible for vision under low light conditions (scotopic vision), while cones are used for color perception (photopic vision) in high lighting where sensors are being tuned to red, green and blue frequencies (RGB) and there, high visual acuity or resolution is achieved.

From the point of view of signal processing, the information is received by photoreceptors and the neuronal network could just be performing mathematical operations combining information via interconnections between the neurons at the different layers, to process the incoming information.

Historically–probably because of the low knowledge about how it works–, multiple extraordinary capabilities have been attributed to the brain such as recovering data or complementing partial information. Visual hyperacuity has long been believed to be related to these brain skills but, however, it is difficult to believe that it can recover data and information which has not been previously captured by the sensors.

As previously mentioned, hyperacuity can be defined as the capacity of the human eye to overcome the resolution determined by the number and size of its sensors [13], [14]. From the point of view of physiology, the Minimum Angle of Resolution (MAR) is established very close to 1 minute of arc for acute vision [15]. Nevertheless, the Vernier acuity test shows how efficiently the eye can resolve parallel lines placed at distances of few seconds of arc [16]. This phenomenon could never be explained if we assume that the human eye operates the same way as current CCD or CMOS sensors, where the maximum resolution is defined by the number and size of pixels in the sensor. These conventional devices work as diffraction limited systems, meaning that the diffraction blur created by the entrance aperture is limited by fitting the most energy of each point from the real scene in each pixel of the sensor. Thus, every single pixel sensor receives independent and uncorrelated information, avoiding various pixels from capturing information from the same point and capturing a focused and sharp image. Normally, in regular sensor arrays, the Rayleigh criterion [17] defines the maximum diffraction that the system can cope with and establishes separation between sensors to resolve contiguous point sources, thereby avoiding overlapping between pixels or sensors. Working under this principle, it is hard to imagine a signal processing technique to achieve visual hyperacuity or enhance the image resolution. The maximum recoverable frequency is limited by the Nyquist sampling theorem (also the maximum spatial resolution), and thus the high-frequency information lost in the process of sensing—or sampling—could never be recovered. In addition, if a resolution increase would be attempted, any interpolation method would involve a low-pass filtering. Nevertheless, it is a matter of fact that the visual hyperacuity is real, so something is missing on this interpretation and it should be a simple scheme, since it is performed by a simple architecture composed only by photoreceptors and the brain's neural network.

Different research studies [14], [18], [19] find an explanation for hyperacuity in the microsaccade movements of the human eye. Related to this, there are also interpretations based on the randomness of these movements discussing the possibility of modeling them by some stochastic process [20], [21]. However, it is rather unlikely that the eye could reconstruct and improve the image resolution using and controlling these small and fast random saccades. In contrast, microsaccade movements may be helpful to refresh sensors and prevent them from saturation, mostly in the object edges, due to their time-integrating behavior [22].

Our central thesis is that the human eye, far from being a diffraction-limited system, could be benefiting from diffraction to achieve hyperacuity and increase image resolution. In fact, the pupil could be acting as an aperture, creating a diffraction pattern or Point Spread Function (PSF), much bigger than a cone or photoreceptor, and projecting a blurred image over the retina. This could be helpful to achieve hyperacuity, since a wider area and multiple sensors are involved in the detection process, creating spatial diversity. The fact that various sensors receive the same information, improves the sensitivity of the system and increases the Signal-to-Noise Ratio (SNR). Therefore, the dynamic range of the sensor is further enhanced. The spreading of light over different sensors because of diffraction could also help to resolve at any point the color information since various RGB-cones are involved in the capturing process, providing not only luminance information but also related to color.

In addition, since diffraction acts as a low-pass filtering process, sharp transitions of the real image (high frequencies) are converted into smooth and slowly varying functions (lower frequencies), enabling the use of a reduced number of sensors for sampling (capturing) the image and allowing a more accurate interpolation if required. In that regard, some authors [23] disclose how super-resolution can be achieved profiting from the spreading of light produced by diffraction. Another article [24] shows how spatial smoothing is totally recoverable in certain conditions and [25] explains how diffraction is used to improve details in seismic imaging. This concept is also applied by [26] in synthetic aperture radar (SAR) imaging, where the azimuth focus is enhanced thanks to the blur created due to high-frequency vibrations of the radar sensors platform.

Multiple factors could be producing a diffraction or blur effect in the eye. First, from an optical theory perspective, light waves are diffracted when passing through an aperture and a lens system, transforming points into blobs and creating a blurred image. The time-integrating behavior of cones [27], [28] added to saccade movements produced in the eye, could also contribute to this blur effect. At last, the photon capturing mechanism of cones produces chemical reactions that creates a coupling effect in adjacent photoreceptors [29]. All these factors could be contributing to project a strongly blurred image over the sensors, even neglecting the Rayleigh criterion. Accepting this fact, the neural network should just solve an inverse problem or deconvolution, which seems not to be so complicated as simple mathematical operations are involved. At this point, the capability of the human eye to solve in an effective way the inverse problem could be questioned, understanding it as the ability to recover a clear and sharp image from a distorted, blurred one.

In our study, we assume that photoreceptors are just power detectors, generating an electrical signal proportional to the power received at a certain interval of time. This seems reasonable, considering the simple architecture of cones, their time-integrating behavior and the real capability to perceive different colors, called chromatic response RGB-cones (sensible to red, green and blue colors) or SML-cones, tuned to capture low, medium and long wavelength frequencies.

## IV. VISUAL HYPERACUITY SIMULATION METHOD

As already explained, the human eye uses a low number of sensors to capture images and, moreover, the retinal image seems to be blurred because of diffraction. We developed a method that intends to improve the final image resolution when the number of available sensors is limited, using a controlled diffraction effect introduced at the beginning of the process, as the human eye does based on our hypothesis. In short, it can be understood as an explanation and simulation of the visual hyperacuity, a diffraction-enhanced system, far from conventional imaging devices, widely known to be diffraction-limited. The proposed method and its performance, both have been tested with the MATLAB software.

The first step of the approach consists in introducing a known diffraction blur using a PSF (low-pass filter, LPF) at the entrance of the system. The resulting blurry image is then captured with a low number of sensors, i.e., spatially subsampled: this would be the input image, the one projected over the sensors and captured by the imaging device. On the next stage the image goes through an interpolation process, increasing the total data points, thus, creating new samples. At the final stage, an inverse filter is applied to remove the blur introduced by the diffraction, which recovers a sharp and focused image, with an enhanced resolution in comparison to the one captured by the sensor. The block diagram of the whole system can be observed in Fig. 1.

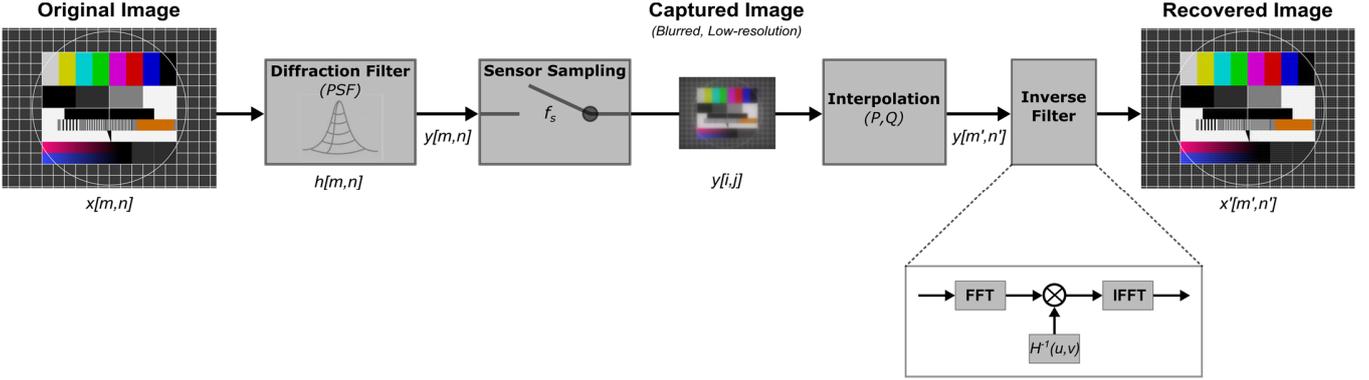

Fig. 1. Block diagram of the visual hyperacuity simulation method: original test image (observed real scene), captured image (input image, blurred and low-resolution) and recovered image (enhanced resolution).

*A. Image Capture: Diffraction and Sensing*

The selected original test image is 1500×2000 pixels in size (see Fig. 2). This image is used as a model of a real scene observed by the imaging system, and as such, it is conceived as a high-resolution image at the entrance of the optical system.

The image projected over a sensor can be modeled by a simple convolution of the real scene and the diffraction pattern or PSF produced by the optical system,

$$y[m,n] = x[m,n] * h[m,n] \qquad (4)$$

where $x$ represents the original test image, $h$ is the PSF and $y$ is considered as the blurry image projected over the sensor. Applying the Fourier Transform, the image acquisition process can be expressed as a simple matrix multiplication as in (5).

$$Y(u,v) = X(u,v) \cdot H(u,v) \qquad (5)$$

The diffraction pattern (PSF) created by a uniformly-illuminated circular aperture is an Airy Disk such as the one displayed in Fig. 3, and its main effect is the introduction of blur in the incoming image, transforming point sources into blobs. The diffraction introduced by the optical system acts as a low-pass filter (LPF) when analyzing in the frequency domain, as previously explained in (5). To simulate the image capturing using a reduced number of sensors, the previously diffracted image is spatially subsampled, performing a decimation by a factor of 10 in each dimension (vertical and horizontal) and reducing data points to 150×200 pixels. It is important to note, that the LPF introduced by the PSF enables to lower the sampling frequency (i.e., use less sensors), since the signal is now limited in a narrower frequency-band. In this sense, we can understand the introduction of diffraction as an anti-aliasing filter, prior to the sampling process, which attenuates considerably the high frequencies of the image. In addition, the subsampling produces spectral truncation, permanently removing some high-frequency spectral components.

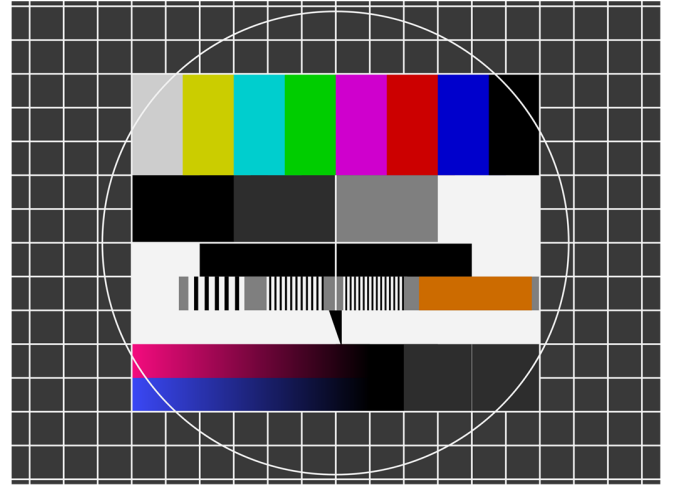

Fig. 2. Original test image used for simulation, reproducing the high-resolution observed real scene (1500×2000 pixels).

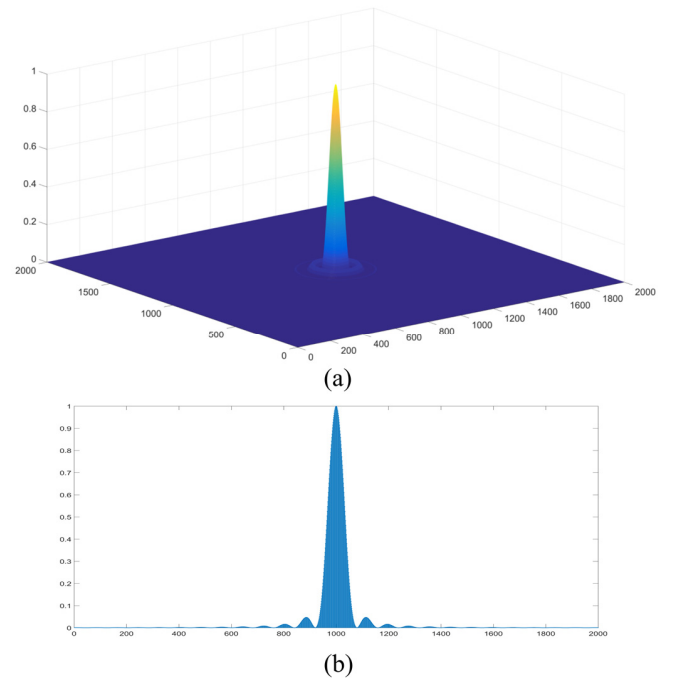

Fig. 3. (a) MATLAB 3D plot of the Point Spread Function. (b) Cross section of the PSF through the central plane.

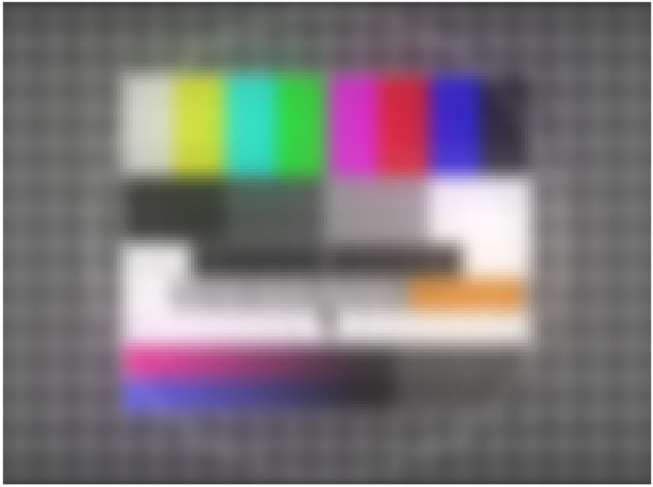

Fig. 4. Image captured by the sensor: blurred and low-resolution (150×200 pixels).

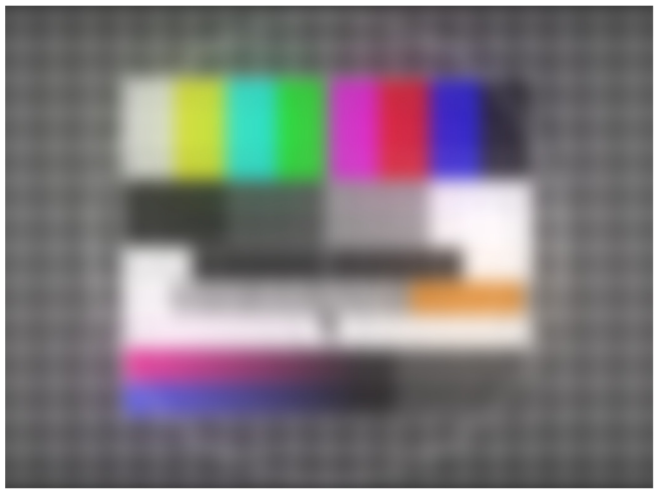

Fig. 5. Interpolated image: blurred and increased data points (1500×2000 pixels).

At this point, we obtain the image considered to be the system input, the one captured by the sensors: a low-resolution image (simulating a low number of sensors) and blurred due to the diffraction introduced by the optical system (see Fig. 4), which renders difficult to appreciate sharp edges and shapes, because of the loss of information at high-frequencies.

## B. Post-Processing: Interpolation and Inverse Filter

To achieve a resolution enhancement, the first step is to increase the image size, and by doing so, new data points are created between the existing samples. For this purpose, 2D interpolation by a factor of 10 has been implemented, using the FFT method. The interpolated image is still blurred (there is no acuity or detail improvement), but the image size is increased to 1500×2000 pixels (observe Fig. 5). The LPF (diffraction) introduced in advance makes possible a more accurate interpolation, since the blurred image presents low-variation and smooth transitions, enabling a more precise estimation of new image data points.

The last stage of the visual hyperacuity simulation method is the application of an inverse filter to remove the blur introduced by the diffraction, obtaining a resolution-enhanced and sharp image. It is worth mentioning that the inverse-problem is ill-posed and thus the inverse filtering process becomes highly unstable, presenting high sensitivity against errors produced by interpolation and noise-amplification problems. However, the LPF effect of the PSF and the subsampling performed in advance, considerably attenuate or even remove some high-frequency components (the most critical in relation to instability and noise-amplification problems), making possible the use of an inverse filter in a straightforward manner.

The inverse filtering is also carried out in the frequency domain, by applying the inverse filter $H^{-1}(u,v)$, to the diffracted, subsampled and subsequently interpolated image as formulated in (6).

$$X'(u,v) = Y(u,v) \cdot H^{-1}(u,v) = Y(u,v) \cdot \frac{1}{H(u,v)} \qquad (6)$$

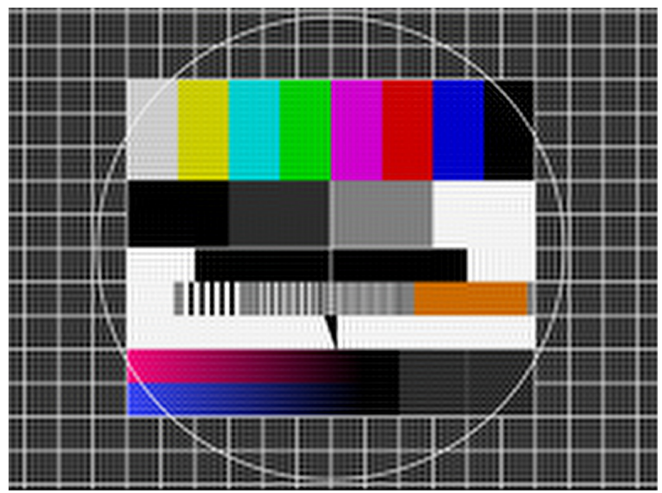

Fig. 6. High-resolution recovered image (1500×2000 pixels), using the visual hyperacuity simulation method.

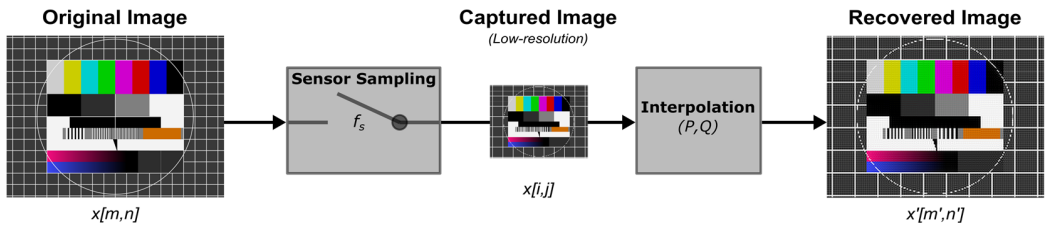

Fig. 7. Block diagram of a diffraction-limited system simulation method: original test image (observed real scene), captured image (input image, low-resolution but diffraction-free) and recovered image (interpolated).

## V. RESULTS

The output image obtained from implementing the inverse filter is shown in Fig. 6. Now edges appear considerably sharper, being possible to recognize and resolve shapes, details, lines and abrupt color transitions, that were not detectable previously. This resolution improvement is produced thanks to the new data points created by interpolation between sensed samples and the reconstruction of the spectral components attenuated in advance by diffraction, but it is not possible to recover the components previously truncated due to the subsampling process. Nevertheless, the most important spectral information of the image is kept in low-frequencies and this loss may not be critical when reconstructing the original image.

In addition, the image has now a size of 1500×2000 pixels, meaning that the resolution is increased by a factor of 10 in each dimension (vertical and horizontal) in comparison to the initial low-resolution image of 150×200 pixels captured by sensors, due to the creation of new data points between samples. However, should be noticed that a slight grid effect is created in the image background because of the FFT interpolation process.

The results obtained by the visual hyperacuity simulation method must be analyzed rigorously if the resolution improvement should be observed. One way of evaluating this resolution enhancement is to compare the proposed method with an identical but diffraction-limited system (without introducing any diffraction): capture the image with the same sensor size of 150×200 pixels and implement an FFT interpolation method, increasing the data points by a factor of 10 in each dimension–for a better understanding the block diagram of this system is depicted in Fig. 7–. The result of reproducing the method with neither diffraction effect nor inverse filtering can be observed in Fig. 8.

Fig. 9 shows side by side the original test image and the images obtained using the proposed method with and without diffraction, for a better visual comparison. Simple inspection and comparison of Fig. 9 (b) and Fig. 9 (c) reveal that thanks to diffraction many fine details can be recovered (grid lines, reconstruction of continuous circumference line, sharp definition of objects keeping their size) and information can as well be detected which is not feasible when diffraction is not introduced in the system. In short, the image shown in Fig. 9 (b) captured in presence of diffraction represents in a more accurate way the original test image represented in Fig. 9 (a), and almost no data are lost. The spreading of light over the sensor produced by diffraction creates spatial diversity and the information of a single point source is also expanded over a wider area, being possible to sense information using a reduced number of sensors, which would be lost in absence of diffraction. Therefore, we can recover information spatially located between sensors and represent it creating new samples: increasing the image resolution.

For the evaluation of the improvement introduced by diffraction in the image-processing method in a more accurate and quantifiable way, it has been computed Peak Signal-to-Noise Ratio (PSNR), an image quality metric which operates at pixel level. The PSNR has been calculated both for the image obtained with the visual hyperacuity simulation method (Fig. 6) and the image obtained without diffraction (Fig. 8), taking the original test image (Fig. 2) as the reference. Table 1 contains the achieved results, showing that better values are obtained for the proposed method, in presence of diffraction.

TABLE I
PEAK SIGNAL-TO-NOISE (PSNR) VALUES FOR THE VISUAL HYPERACUITY SIMULATION METHOD AND DIFFRACTION-LIMITED METHOD.

| Image | PSNR (dB) |
|---|---|
| Visual Hyperacuity (Fig. 6) | 16.30 |
| Diffraction-limited (Fig. 8) | 14.28 |

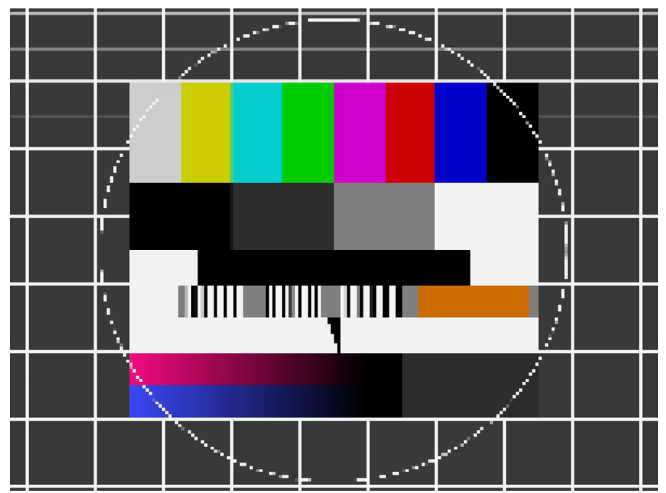

Fig. 8. Recovered image (1500×2000 pixels) simulating a diffraction-limited system and interpolation.

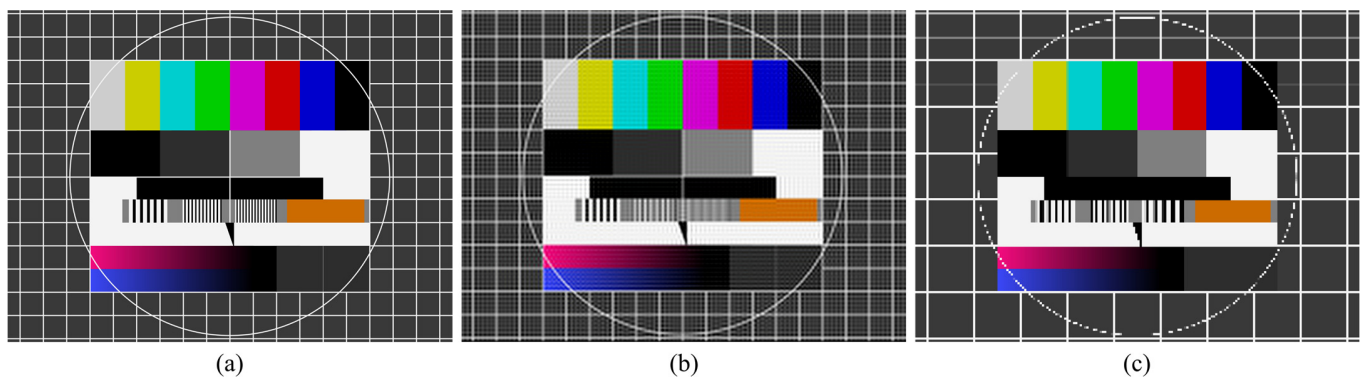

Fig. 9. Visual comparison of (a) original test image, (b) visual hyperacuity simulation method (diffraction) and (c) diffraction-limited simulation method (no-diffraction).

## VI. CONCLUSIONS

This article can be understood as an explanation for the human eye visual hyperacuity, since it has been demonstrated that the introduction of a controlled diffraction pattern could play a key role on surpassing the resolution defined by the number and size of sensors of a given system, also breaking the paradigm of the number of pixel sensors needed to capture an image and its final resolution. Therefore, the human visual system can be understood as a diffraction-enhanced system, far from conventional CCD or CMOS imaging systems broadly known as diffraction-limited.

From the standpoint of sensors, diffraction creates spatial diversity due to the spreading of light over the surface where sensors are located, improving the system sensitivity and increasing the SNR, since multiple sensors receive information from the same point source. In addition, it makes possible to capture or sense fine details that would be spatially located between detectors, which cannot be detected in absence of diffraction, when using a reduced number of sensors. Moreover, diffraction can be understood as an anti-aliasing filter which relaxes significantly signal sampling requirements, allowing the use of a reduced number of sensors for capturing the image. Spectral components truncated because of the subsampling cannot be recovered, but they seem not to be critical, since the most important components are kept in low frequencies, and most of the fine details in the image are resolved.

Yet, it is worth to mention that the inverse problem comes to be ill-posed and the system could show stability problems and be quite sensitive to small errors generated on account to interpolation. This instability can also be a serious issue in presence of noise, which involves using inverse functions or methods to minimize problems associated to noise amplification. Nevertheless, the attenuation and elimination of some high-frequency spectral components produced thanks to the diffraction and subsampling, reduces considerably problems related to stability and noise-amplification, allowing the use of a direct inverse filter. In any case, the choice of the most appropriate method to solve the inverse problem could be affected by diverse factors as the nature of the phenomenon of diffraction, the source of noise, or the required resolution for the specific application.

As a final observation, due to the possibility of obtaining high-resolution images out of a low number of sensors, it would be possible to increase the surface area of each pixel sensor (with no need of increasing the total sensor size) and, this way, improve the dynamic range of the sensor and the total SNR of the system. This concept becomes interesting when vision under poor lighting conditions is required, as real improvement of the sensitivity occurs, unlike digital ISO stands on digital cameras, where both signal and noise are amplified.